# Supporting Assessment of Novelty of Design Problems Using Concept of Problem SAPPhIRE

Sanjay Singh[1[0000-0001-8943-4208]] and Amaresh Chakrabarti[1[0000-0002-1809-1831]]

[1] Indian Institute of Science, Bangalore, India
sanjaysingh@iisc.ac.in

**Abstract.** This paper proposes a framework for assessing the novelty of design problems using the SAPPhIRE model of causality. SAPPhIRE denotes different abstraction levels where S stands for State change, A stands for Action, P stands for Parts, Ph stands for Physical Phenomena, I stands for Input, R stands for oRgan and E stands for Physical Effect. The novelty of a problem is measured as its minimum distance from the problems in a reference problem database. The distance is calculated by comparing the current problem and each reference past problem at the various levels of abstraction in the SAPPhIRE ontology. The basis for comparison is textual similarity. To demonstrate the applicability of the proposed framework, The 'current' set of problems associated with an artifact, as collected from its stakeholders, were compared with the 'past' set of problems, as collected from patents and other web sources, to assess the novelty of the 'current' set. This approach is aimed at providing a better understanding of the degree of novelty of any given set of current problems by comparing them to similar problems available from historical records. By applying such approaches, organizations could effectively prioritize and address emerging problems based on their relative novelty, with positive ramifications on problem-solving and decision-making. Since manual assessment, the current mode of such assessments as reported in the literature, is a tedious process, to reduce time complexity and to afford better applicability for larger sets of problem statements, an automated assessment is proposed and used in this paper.

**Keywords:** Creativity, Novelty, SAPPhIRE.

## 1 Introduction

In the realm of engineering and product development, creativity and innovation play pivotal roles in shaping advancements and meeting evolving consumer demands. Chakrabarti [1] considered engineering design as a central part of the product development process, and it is distinguished from other aspects of engineering by its creative aspects, whereby novel products are conceived. The ability to identify and assess the novelty of design problems is crucial for fostering creativity within the design process. This research focuses on exploring methodologies and frameworks for effective evaluation of novelty of design problems, aiming to enhance the creative potential of engineering design. In today's competitive market, where differentiation and innovation are key drivers of success, businesses are constantly challenged to develop products that not only meet functional requirements but also captivate consumers with fresh and unique attributes. Understanding the novelty of design problems is essential as it lays the groundwork for generating innovative solutions that can address emerging market needs and technological advancements.

By deciphering the degree of novelty of a design problem, engineers and designers could strategically allocate resources, focus efforts on high novelty problems, and explore unconventional solutions that could push the boundaries of traditional thinking.

This paper examines various approaches from the literature on the novelty of design solutions and investigates how one can be applied to effectively assess the novelty of design problems, offering valuable perspectives for researchers, practitioners, and industry stakeholders involved in the pursuit of innovative product development.

## 2 Literature review

### 2.1 Novelty

Novelty resembles unusualness or unexpectedness [4]. According to Sarkar [5], novelty happens when an agent generates an outcome without replicating any existing outcome(s). Novel means "new and original, not like anything seen before" and novelty is the "quality of being new and unusual and something that has not been experienced before, and so is interesting" [6]. Shah et al. [5] and Lopez-Mesa et. al [7] used "infrequency" as a measure of novelty. "Nonobviousness" is used as a measure while assessing novelty in patent documents [8]. Novelty may also be defined with reference, either to the previous ideas of the individual concerned (P-Novelty, P for Psychological) or to the whole of human history (H- Novelty, H for Historical) [9].

The importance of novelty, as argued by Sarkar [5], is that it helps determine a design's newness and patentability, serves as a criterion for comparing a designer's capability, and ascertains the potential market of a product.

### 2.2 Novelty Assessment Methods

Amabile [12] suggests the use of experts to identify what is "creative". Jansson and Smith [13] measure the originality ("O" score—a proxy for novelty) of an idea as:

$$O = 1 - \frac{n}{m} \quad (1)$$

where n is the number of similar ideas in a design session and m is the total number of ideas in a design session.

Shah et al. [5] improved the above metric by including weights for different requirements and design stages. Due to its simplicity, this metric has gained popularity and usage in academia. Lopez-Mesa et. al [7] developed two methods for measuring the novelty of design alternatives. Each design alternative is classified in terms of its action function, structure, and detail. Sarkar and Chakrabarti [2, 14] developed a method for assessing the qualitative relative degree of novelty of an engineering product using the constructs of the product. The various degrees of novelty are very-high, high, medium, or low. According to the authors, the method employs function–behavior–structure (FBS) and SAPPhIRE models together; the FBS model is used first for determining novelty and then the SAPPhIRE model is used to assess the relative degree of novelty. Grace et al. [15] model a class of products as vectors using the K-means clustering algorithm; novelty is calculated as the Euclidean distances among these. Fu et al. [21] construct a Bayesian Network of patents using the cosine similarities among these. Gosnell and Miller [22] assign 5 of 36 adjectives to the design solutions and calculate the novelty scores of these as the average similarity to the word— "innovation" [22].

The above measures fall into two categories: frequency-based and distance-based measures. There are some fundamental issues with the usage of frequency-based measures [16]. Siddharth et. al. [16] demonstrated how the novelty of a solution depends on the cardinality of a solution set in the existing literature. Another limitation is that frequency-based measures do not explain how solutions are compared against each other. Frequency-based measures might overlook nuanced differences between similar items (since they focus purely on how often a feature occurs, rather than how different it is from other features). For example, when assessing a problem against a set of existing problems, since only one problem is considered, frequency becomes irrelevant to the evaluation process. A typical evaluation of a solution should be with respect to a common reference and should not depend on the number of solutions as is in the case of Shah et. al [4]. It is acknowledged that the frequency-based measure does not compute actual novelty [4]; rather, it only provides feedback in the form of presence and chance of novelty to designers [23]. However, the distance-based measures do not include the number of solutions and measure novelty with respect to a common reference which is more appropriate than other measures. From an engineering design standpoint, a priori, novelty means "new" or "original" and is measured relative to the entirety of human history [17]. In practice, such a reference could be a design session (e.g., a competition), a physical boundary (e.g., a city), the prior knowledge of the judge, or a product database (e.g., US patent database). The most practical, and reliable approach would be to use a comprehensive product database as the reference to measure its "distance" from the solution [18].

## 2.3 SAPPhIRE model of causality

The SAPPhIRE model of causality was developed by Chakrabarti et al. [11]; Fig. 1, to explain the causality of natural and engineered systems. The model gets its name from the highlighted letters of its constructs: **state change, action, parts, phenomenon, input, organs, and effect**. Note that the term effect is used to collectively mean a combination of physical laws and effects. The definitions of the constructs have been rephrased in Srinivasan and Chakrabarti [10] to provide greater clarity in understanding the constructs and hence usage of the model.

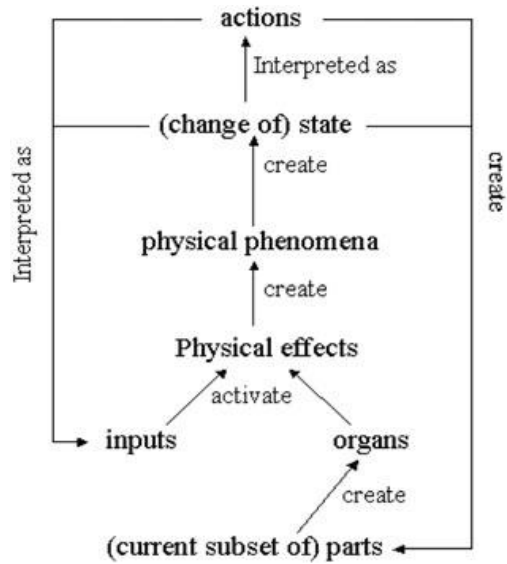

**Fig. 1.** SAPPhIRE Model of Causality [11]

### 2.4 Summary

Sarkar and Chakrabarti [11] use a descriptive model of causality (SAPPhIRE) as the parameters for which distances are assessed against designs or products used as reference. It exhaustively covers (a) the description of the solution; (b) a database of products, ideas, and concepts; and (c) a method for comparing the solution against the database. Therefore, an additional interpretation is not required. Moreover, a SAPPhIRE model is not domain-specific, unlike in the Euclidean distance-based measures. Considering these advantages, the SAPPhIRE-based measure seems to be a suitable one to use and has been used in this work.

## 3 Research Gap and Question

Several methods for assessing the novelty of products exist in the literature. Task clarification stage is the earliest of the stages that can accommodate changes that are most effective and least expensive. At the earliest of task clarification stage where one is dealing with the problems, no one seems to have assessed them for their novelty. On the other hand, the overall novelty of a design solution is influenced by both the novelty of the problems addressed, and the novelty of the solutions developed to address these problems. Hence the focus of this work is on assessing the novelty of design problems. This leads to the research question for our study:

***Research Question:*** How can the novelty of design problems be assessed?

## 4 Research Methodology

To address the research question, a preliminary study was conducted on a design case to assess the novelty of design problems associated with the case. The proposed methodology (Figure 2) involves the following eight steps: (1) collecting design problems in the same context available in patents or online searches and their classification as 'past problems'; and (2) collecting design problems using a questionnaire or survey those currently faced by stakeholders in that context as 'current problems''; (3) describing each of the past and current problems so collected using the SAPPhIRE model of causality, henceforth called Problem SAPPhIREs [19]; (4) comparing each natural language description of past and current problem SAPPhIRE in these two distinct databases with one another for similarity at the action level of SAPPhIRE (the comparison is facilitated using BERT semantic similarity with a pre-trained database); (5) For those problem-SAPPhIREs for which action level of SAPPhIRE matches, performing sentence similarity, along with word-to-word similarity, between each pair of problem-SAPPHIREs by converting their words into word-vectors and calculating the cosine distance between them, at each, remaining level of SAPPhIRE abstraction [16]; and (6) Calculating Similarity score for each level of SAPPhIRE abstraction, where Dissimilarity (Novelty) scores for each pairwise comparison is determined using the following equation: Novelty = (1 – Similarity) [16]; (7) calculating the overall novelty score of the given, current problem, in comparison to the previous problem under consideration, by averaging these pairwise novelty scores; (8) interpreting the novelty score qualitatively using the following ranges:
0 to 0.3 as low novelty; 0.3 to 0.7 as medium novelty; and 0.7 to 1 as high novelty.
Using the pre-trained dataset and a web-based tool, some of the above processes (steps 4 and 5) are automated.

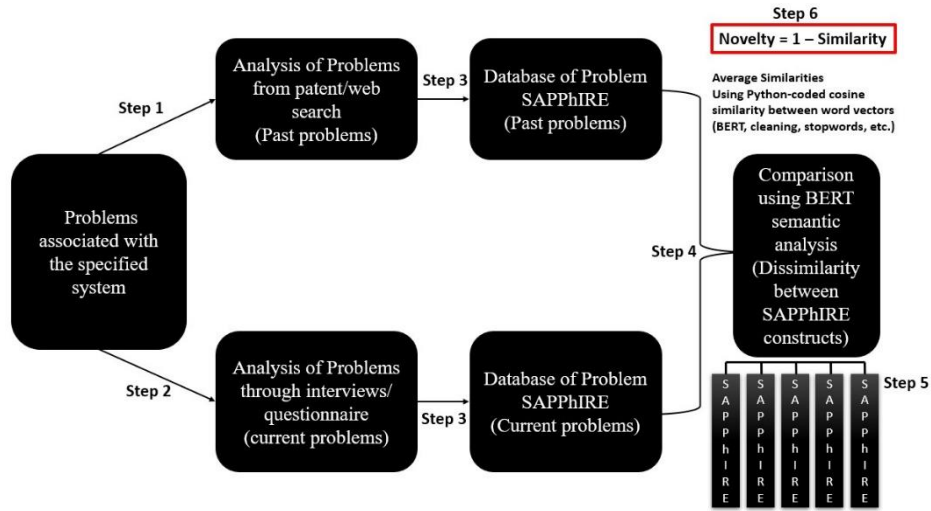

**Fig. 2.** Framework for the assessment of novelty of design problems

Among the current problems that were available, one problem was found to be of high novelty. The case study below illustrates the process.

## 5   Case study (Electric Kettle) demonstrating the application of the framework

We decided to use an electric kettle as the system for the problem analysis. Different parts of the electric kettle with their functions are shown in Fig. 3. Utilizing online patent libraries and patent searches, a comprehensive examination of electric kettle patents during the last 100 years was conducted. Data on patents was extracted both manually and with the use of a Python program called Scrapy. After that, useful data in the form of past design problems are extracted, and using these, a database of past problems for this system is developed. In Table 1, a few of the problems collected from the past problem database are enlisted.

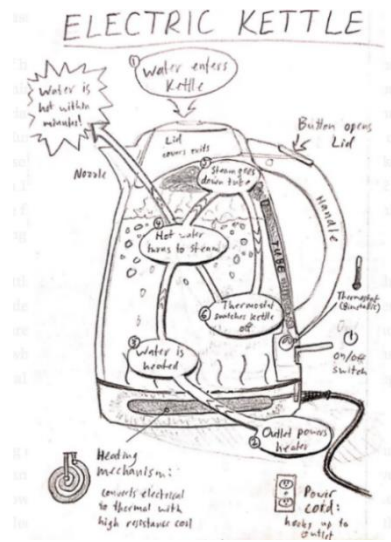

**Table 1.** A few of the design problems collected from patent and web sources

**Fig. 3.** Functionality and part description of Electric Kettle [20]

| S.No. | Past Design Problems |
|---|---|
| 1. | Avoiding the overheating of the liquid present in the kettle. |
| 2. | Providing safety to the users from hot steam coming out. |
| 3. | Less heat transfer out of the kettle. |
| 4. | Spilling of the liquid which decreases the efficiency of the kettle. |
| 5. | Temperature and boiling control to minimize inefficient heating. |
| 6. | Tough to clean. |

Moving forward, we collected a set of current design problems, as mentioned by the stakeholders, using a survey questionnaire. Those problems were stored in a database of current problems. A few of these problems are enlisted in Table 2.

**Table 2.** A few of the design problems collected from the survey questionnaire

| S.No. | Current Design Problems |
|---|---|
| 1. | Water is not getting heated. |
| 2. | The user's hands are getting burnt due to the steam coming out. |

| 3. | Overheating takes place since there is no temperature monitoring available. |
|---|---|
| 4. | When water overboils it spills out. |
| 5. | Cleaning is tough inside intricate areas of the kettle base. |
| 6. | Underheating of the coil element due to corroded coil. |
| 7. | Localized roughness in the kettle base due to longer scrubbing results in local heating/boiling. |
| 8. | The conical shape makes it difficult to wash |
| 9. | Heating does not stop due to a problem with the automatic shutoff |

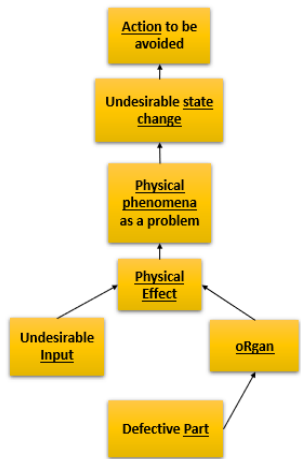

**Fig. 4.** Problem SAPPhIRE model [19]

The past problems are analyzed and transformed into their Problem-SAPPhIREs using the Problem SAPPhIRE model [19]. The current problems are also analyzed and transformed into their Problem SAPPhIREs.

The problem of "spilling of liquid" was selected as the example case for novelty assessment, based on its high frequency in patents. Corresponding Problem SAPPhIREs 1 and 2 (Fig. 5), based upon all types of liquid spilling within an electric kettle, are chosen from the database of past problems as reference past problems for this case. Similarly, Problem-SAPPhIREs 3, 4, and 5 (Fig. 6) are chosen as current problems for this case. In these problems, their action level description includes "spilling of liquid" as the action.

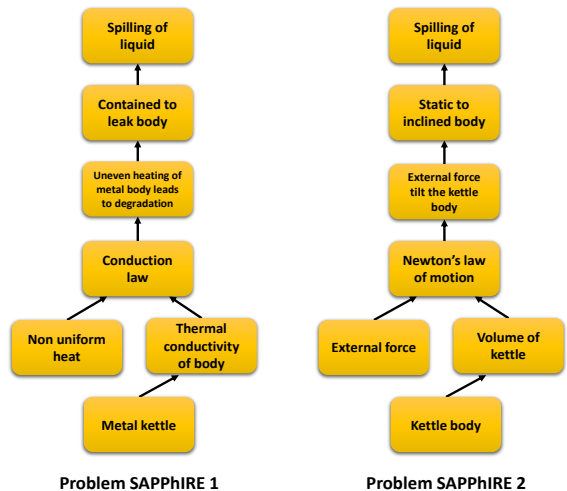

**Fig. 5.** Problem SAPPhIREs of "Spilling of liquid" for past problems taken from patents

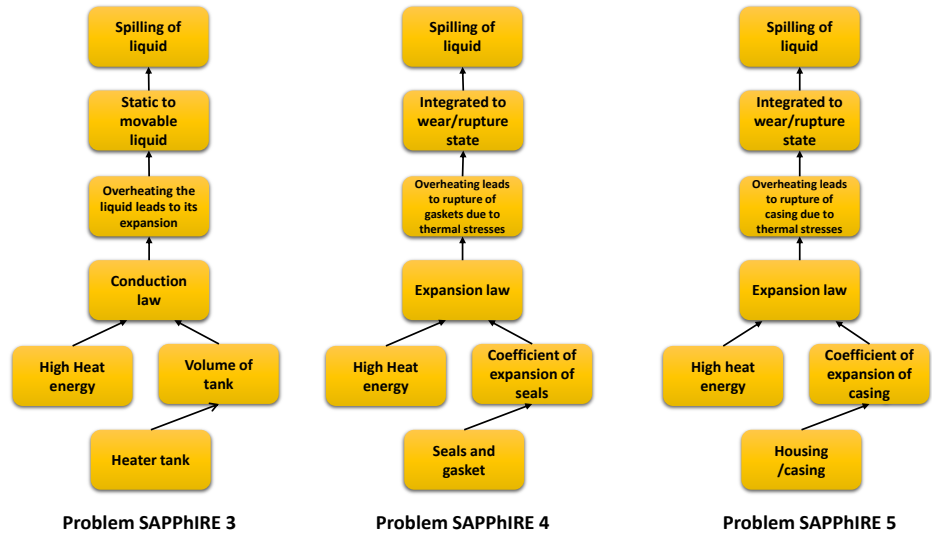

**Fig. 6.** Problem SAPPhIREs of "Spilling of liquid" for current problems from the questionnaire

## 6      Results

The arrows in Fig.7 and Fig. 8 indicate comparison of PS1 and PS2 with PS3, PS4, and PS5 (PS- Problem SAPPhIRE) respectively.

When comparing PS1 and PS3, the Sentence at the State change level "Contained to leak body" is assessed for semantic similarity with "static to movable liquid", and the result is 0.314, indicating that both are quite dissimilar. Following the comparison of the state change level, the semantic similarity of Problem SAPPhIRE 1 is performed with each of Problem SAPPhIREs 3, 4, and 5, at the Phenomena, Effect, Input, oRgan, and Parts levels. The novelty scores are enlisted in Table 3.

Similar activities are also performed for the past Problem SAPPhIRE 2, with Problem SAPPhIRE 3, 4, and 5 each. Table 4 tabulates the novelty scores for these comparisons. Scores range from 0 (zero novelty) to 1 (high novelty), with real values contributing to the final determination of the qualitative nature of the novelty.

The average novelty score for a given problem is determined after the novelty scores for each abstraction level are computed.

The classification of low, medium, and high novelty is then done using these scores. Problem 5 is regarded as problem with high novelty because it has a maximum novelty score of 0.7 in both the scenarios (in comparison with Problem 1 and 2). With scores ranging from 0.3 to 0.7, problems 3 and 4 are classified of medium novelty.

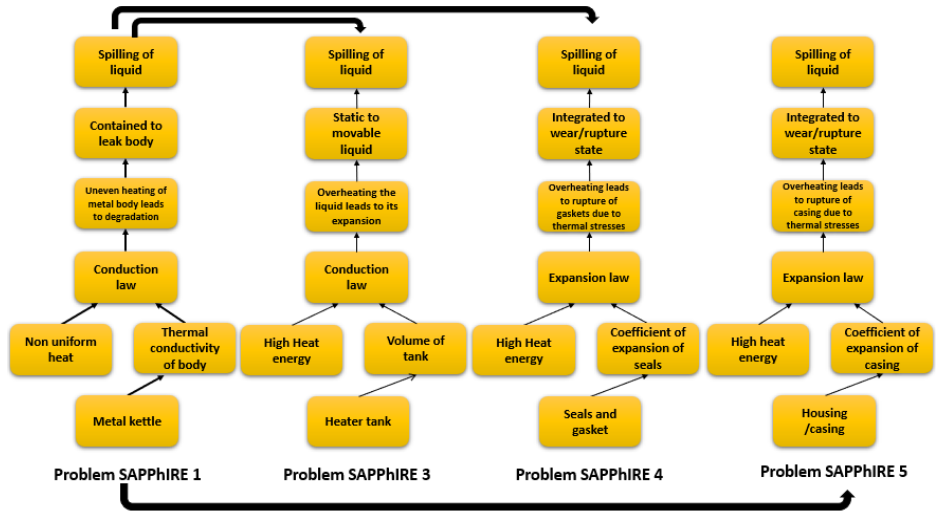

**Fig. 7.** Comparison of Problem SAPPhIRE 1 with 3, 4, and 5

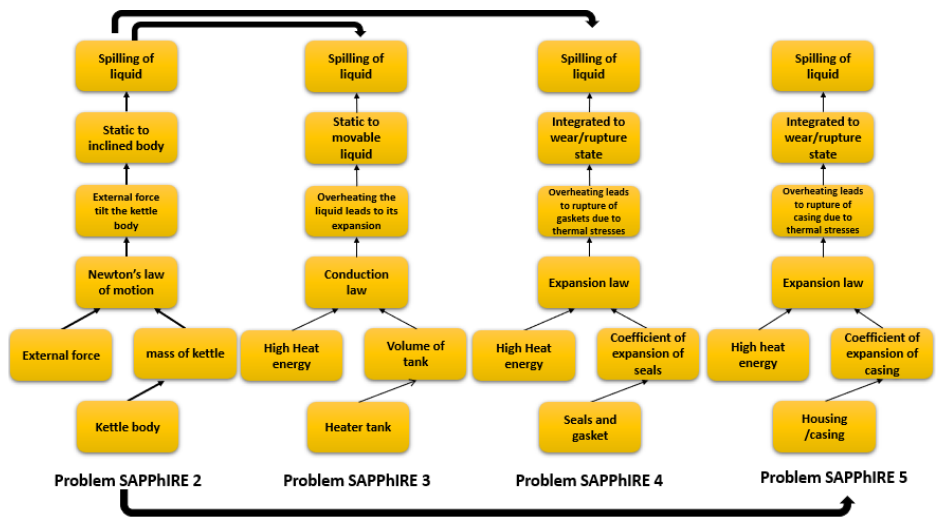

**Fig. 8.** Comparison of Problem SAPPhIRE 2 with 3, 4, and 5

Table 3.  Novelty score and its interpretation after comparing PS1 with PS3, PS4, and PS5

| Constructs/ Comparison pair | PS1-PS3 | PS1-PS4 | PS1-PS5 |
|---|---|---|---|
| **A**ction | 0 | 0 | 0 |
| **St**ate Change | 0.686 | 0.679 | 0.679 |
| **Ph**enomena | 0.519 | 0.587 | 0.68 |
| **E**ffect | 0 | 0.629 | 0.629 |
| **I**nput | 0.699 | 0.699 | 0.699 |
| o**R**gan | 0.796 | 0.628 | 0.725 |
| **P**arts | 0.613 | 0.694 | 0.822 |
| Avg. Novelty | 0.55 | 0.65 | 0.7 |
| Cumulative Decision on Novelty | Medium Novelty | Medium Novelty | High Novelty |

Table 4.  Novelty score and its interpretation after comparing PS2 with PS3, PS4, and PS5

| Constructs/ Comparison pair | PS2-PS3 | PS2-PS4 | PS2-PS5 |
|---|---|---|---|
| **A**ction | 0 | 0 | 0 |
| **St**ate Change | 0.506 | 0.553 | 0.553 |
| **Ph**enomena | 0.625 | 0.779 | 0.799 |
| **E**ffect | 0.662 | 0.755 | 0.755 |
| **I**nput | 0.763 | 0.763 | 0.763 |
| o**R**gan | 0.535 | 0.579 | 0.588 |
| **P**arts | 0.644 | 0.643 | 0.731 |
| Avg. Novelty | 0.62 | 0.68 | 0.7 |
| Cumulative Decision on Novelty | Medium Novelty | Medium Novelty | High Novelty |

## 7    Discussion and Conclusions

In the context of applying current novelty assessment methods for design problems, we encountered a scenario where the historical data has only two instances of "spilling of liquid" while the current data featured three analogous cases. To ascertain the novelty within this context, we proposed a structured assessment approach. Initially, we conducted comparisons of Problem 1 and Problem 2 (past problems) with each of Problems 3, 4, and 5 (current problems) to assess which among these three current problems exhibited the highest degree of novelty. Following this evaluation, we proceeded to assess the novelty values of these three problems relative to one another. For each current problem, its lowest novelty (in comparison to all the past problems) was ascertained. These lowest values were then compared with one another to order these current problems for their relative novelty with respect to one another. Problem 5 was found to be of the highest novelty among the three current problems assessed. Despite its novelty value not being significantly higher, Problem 5 stood out as the most novel compared to Problem 3 and Problem 4.

The initial intent was to check whether the design problems can be assessed for their novelty. Our study indicates this to be viable by comparing current design problems with the relevant, past ones. This approach enables a clearer understanding of the uniqueness and distinctiveness of current problems compared to those from historical records. Reasoning for generation and validation of Problem SAPPhIRE is a part of future work planned. What we have shown so far, however, is limited to only two sets of comparisons. A complete analysis will require making an exhaustive set of comparisons (among all currently available problems with all relevant problems from the past) and use the least value of novelty for each current problems to order them in terms of their relative novelty against one another. Further, the current framework needs to be validated against the intuitive notion of experts. Another limitation of the current work is that the problems compared are all from matching contexts, while similar problems may have been in existence also from different contexts. The novelty, therefore, needs to be assessed against similar problems across different contexts, which is also a part of our future work. Another opportunity is the development of additional metrics for assessing the value of the design problems (e.g. feasibility, impact, and complexity) that could provide a more holistic view of the creative potential of design problems. Areas like academic research, startups, societal challenges, market innovation, and R&D will be the direct beneficiaries of the proposed method.